\documentclass[letterpaper, 10 pt, conference]{ieeeconf}  

\IEEEoverridecommandlockouts                              

\usepackage{graphics} 
\usepackage{epsfig} 
\usepackage{mathptmx} 
\usepackage{amsmath} 
\usepackage{amssymb}  
\usepackage{graphicx}
\usepackage{bm}
\usepackage{eucal}
\usepackage{todonotes}[inline]  

\usepackage{float}
\usepackage[ruled,vlined]{algorithm2e}
\usepackage{pgfplotstable}
\usepackage{pgfplots}
\usepackage{tikz}
\usetikzlibrary[pgfplots.groupplots]
\usepgfplotslibrary{fillbetween}
\pgfplotsset{compat=1.5}
\usepackage{subcaption}
\usepackage{caption}
\usepackage{xcolor}
\usepackage{enumerate}
\usepackage{makecell}
\usepackage{colortbl}

\usepackage{graphicx}
\usepackage{tabularx,ragged2e,booktabs}
\usepackage{url}
\usepackage{siunitx}
\usepackage{rotating}
\usepackage{amssymb}
\usepackage{tabularx}
\usepackage{pgfplots}
\usepackage{csvsimple}
\usepackage{multirow}
\usepackage{cite}
\usepackage{float}
\usepackage{graphicx}
\usepackage{booktabs}

\DeclareMathOperator*{\argmax}{arg\,max}

\title{\LARGE \bf
Comparison of Information-Gain Criteria for Action Selection
}

\author {Prajval Kumar Murali and Mohsen Kaboli
\thanks{P.K.Murali and M.Kaboli are with the BMW Group, Munich Germany.
e-mail: name.surname@bmwgroup.com}%
\thanks{P.K. Murali is with the University of Glasgow, Scotland}%
\thanks{M. Kaboli is with the Donders Institute for Brain and Cognition, Radboud University, Netherlands }%
}
\begin{document}

\maketitle
\thispagestyle{empty}
\pagestyle{empty}

\begin{abstract}
Accurate object pose estimation using multi-modal perception such as visual and tactile sensing have been used for autonomous robotic manipulators in literature. Due to variation in density of visual and tactile data, a novel probabilistic Bayesian filter-based approach termed translation-invariant Quaternion filter (TIQF) is proposed for pose estimation using point cloud registration. Active tactile data collection is preferred by reasoning over multiple potential actions for maximal expected information gain as tactile data collection is time consuming. 
In this paper, we empirically evaluate various information gain criteria for action selection in the context of object pose estimation. We demonstrate the adaptability and effectiveness of our proposed TIQF pose estimation approach with various information gain criteria. We find similar performance in terms of pose accuracy with sparse measurements ($<15$ points) across all the selected criteria. Furthermore, we explore the use of uncommon information theoretic criteria in the robotics domain for action selection.


\end{abstract}

\section{INTRODUCTION}
\label{sec:introduction}
        

Accurate estimation of object pose (translation and rotation) is crucial for autonomous robots to grasp and manipulate objects in an unstructured environment. Even small inaccuracies in the belief of the object pose can generate incorrect grasp configurations and lead to failures in manipulation tasks~\cite{Qiang-TRO-2020}. 
Strategies based on vision sensors are commonly used for estimating the pose of the object, but there is residual uncertainty in the estimated pose due to incorrect calibration of the sensors, environmental conditions (occlusions, presence of extreme light, and low visibility conditions), and object properties (transparent, specular, reflective). Tactile sensors in combination with robot proprioception provides high fidelity local measurements regarding object pose. However, mapping entire objects using tactile sensors is highly inefficient and time-consuming which necessitates the use of active data collection for object pose estimation. Furthermore, due to extremely low sparsity of the tactile data, novel techniques are required for performing pose estimation.


\begin{figure}[t!]
\centering
   \includegraphics[width = \columnwidth, height = 7cm, trim=1cm 0.1cm 0.5cm 0.1cm, clip=true]{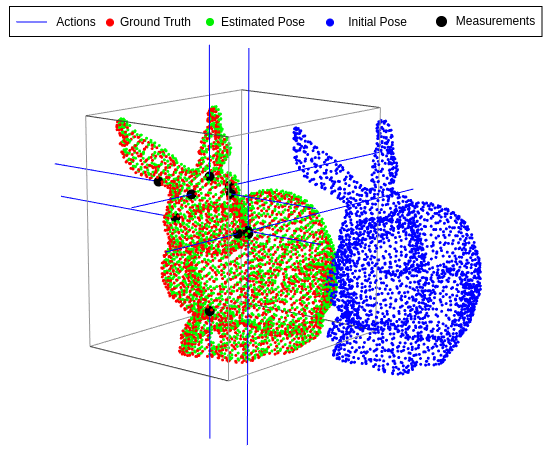}
   \caption{Active pose estimation depicted on the simulated Bunny dataset showing an initial pose (in blue), the estimated pose after the filter-based registration (in green), ground truth pose (in red) and measurements that are actively selected (in black) by performing action selection (blue lines). }
   \label{fig:figure1}
\end{figure}

Typical batch registration methods for pose estimation such as ICP or its variants~\cite{pomerleau2013comparing} have low performance when sparse data is available that arrive sequentially as is the case with tactile measurements~\cite{glozman2001surface}. 
Hence filter-based approaches are generally preferred for sequential data~\cite{arun2019registration,petrovskaya2011global}.
Tactile measurements are inherently sparsely distributed and a probabilistic method was proposed in~\cite{arun2019registration} to perform registration given sparse point cloud, surface normal measurements and the geometric model of the object.
While tactile data can be collected in a randomised manner or driven by a human-teleoperator, active touch strategies which allows for autonomous data collection and reduction of redundant data collection are required~\cite{kaboli2019tactile}.
Several works have used information gain metric based on the uncertainty of the object’s pose to determine the next best touching action to localise the object~\cite{kaboli2017tactile}.
While in literature, the next best action selection is based on expected information gain via metrics such as Shannon entropy~\cite{kaboli2019tactile}, Kullback–Leibler divergence~\cite{petrovskaya2016active}, mutual information~\cite{kaboli2017tactile} and so on, a number of other related information theoretic metrics remain to be explored in the robotic domain. 

\textbf{Contribution: }In this article, we empirically evaluate various information theoretic criteria for selection the next best action in the context of tactile-based localisation. We use our novel probabilistic translation-invariant Quaternion filter (TIQF) for pose estimation~\cite{murali2021active}. TIQF decouples the estimation of rotation and translation and estimates rotation using a linear Kalman filter. 
We empirically evaluate the following criteria: (a) Kullback-Liebler divergence, (b) R\'enyi divergence, (c) Wasserstein distance, (d) Fisher information metric and (e) Bhattacharya distance. Due to our novel TIQF formulation enforcing multivariate Gaussian distributions for prior and posterior rotation estimate, all criteria have closed form solutions. This further enables to exhaustively search for the next best action with marginal computation time overhead.

\section{METHODS}
\label{sec:methods}
\subsection{Translation-Invariant Quaternion Filter (TIQF)}
\label{sec:tiqf}
To solve the point cloud registration problem, we design a linear translation-invariant quaternion filter (TIQF)~\cite{murali2021active}. 
The point cloud registration problem given known correspondences can be formalised as follows:
\begin{equation}
     \mathbf{s}_i = \mathbf{R}\mathbf{o}_i + \mathbf{t} \quad i = 1, \dots N \quad ,
     \label{eq:generativemodel}
 \end{equation}
 where $\mathbf{s}_i \in \mathbb{R}^3$ are points belonging to the scene cloud $\mathcal{S}$ drawn from sensor measurements and $\mathbf{o}_i \in \mathbb{R}^3$ are the corresponding points belonging to the model cloud $\mathcal{O}$.
 Rotation and translation are defined as $\mathbf{R} \in SO(3)$ and  $\mathbf{t} \in \mathbb{R}^3$ which are unknown and need to be computed in order to align $\mathbf{o}_i$ with $\mathbf{s}_i$. We decouple the rotation and translation estimation as translation can be trivially computed once rotation is known~\cite{horn1987closed}.
 Given a pair of correspondences $(\mathbf{s}_i, \mathbf{o}_i)$ and $(\mathbf{s}_j, \mathbf{o}_j)$, we define $\mathbf{s}_{ji} = \mathbf{s}_{j} - \mathbf{s}_{i}$ and $\mathbf{o}_{ji} = \mathbf{o}_{j} - \mathbf{o}_{i}$. From Equation~\eqref{eq:generativemodel} we have:
\begin{align}
    \mathbf{s}_j - \mathbf{s}_i &= (\mathbf{R}\mathbf{o}_j + \mathbf{t}) - (\mathbf{R}\mathbf{o}_i + \mathbf{t}) \quad ,\\
    \mathbf{s}_{ji} &= \mathbf{R}\mathbf{o}_{ji}  \quad . 
    \label{eq:trans_invariance}
\end{align}
Equation \eqref{eq:trans_invariance} is independent of $\mathbf{t}$ and once rotation ${\mathbf{R}}$ is estimated, the translation ${\mathbf{t}}$ can be obtained in closed form from Equation \eqref{eq:generativemodel}.




We cast the rotation estimation problem into a Bayesian estimation framework and estimate it using a Kalman filter.
We define the rotation as the state $\mathbf{x}$ of our filter. Objects are assumed to be fixed, therefore the true state $\mathbf{x}$ is static and does not change over time.
To leverage the insights from Equation \eqref{eq:trans_invariance}, we formulate a linear measurement model as in~\cite{arun2019registration}. We can rewrite \eqref{eq:trans_invariance} as: 
\begin{equation}
    \widetilde{\mathbf{s}}_{ji} = \mathbf{x} \odot \widetilde{\mathbf{o}}_{ji} \odot \mathbf{x}^{*} \quad . 
    \label{eq:quat_objective}
\end{equation}
Since $\mathbf{x}$ is a unit quaternion, we use $\sqrt{\mathbf{x}\odot \mathbf{x}^{*}} = ||\mathbf{x}||=1$ to get
\begin{align}
    \widetilde{\mathbf{s}}_{ji}\odot \mathbf{x} &= \mathbf{x} \odot \widetilde{\mathbf{o}}_{ji} \\
    \widetilde{\mathbf{s}}_{ji}\odot \mathbf{x} &- \mathbf{x} \odot \widetilde{\mathbf{o}}_{ji} = 0 \quad .
    \label{eq:quat_objective_2}
\end{align}
We can further rewrite~\eqref{eq:quat_objective_2} using the matrix notation of quaternion multiplication as: 
\begin{align}
    \begin{bmatrix}
        0 & -\mathbf{s}{ji}^T \\
        \mathbf{s}_{ji} & \mathbf{s}_{ji}^{\times}
    \end{bmatrix}\mathbf{x} -  \begin{bmatrix}
        0 & -\mathbf{o}{ji}^T \\
        \mathbf{o}_{ji} & -\mathbf{o}_{ji}^{\times}
    \end{bmatrix} \mathbf{x} = \mathbf{0} \\
    \begin{bmatrix}
        0 & -(\mathbf{s}_{ji} - \mathbf{o}_{ij})^T \\
        (\mathbf{s}_{ji} - \mathbf{o}_{ji}) & (\mathbf{s}_j + \mathbf{s}_i + \mathbf{o}_j + \mathbf{o}_i)^{\times}
        \end{bmatrix}_{4 \times 4} \mathbf{x} &= \mathbf{0} 
        \label{eq:expected_measurement}
\end{align}
Note that $\mathbf{x}$ lies in the null space of $\mathbf{H}_t$.
Similar to~\cite{arun2019registration}, a \textit{pseudo measurement model} for the Kalman filter is defined:
\begin{align}
    \mathbf{H}_t \mathbf{x} &= \mathbf{z}^h \quad .
    \label{eq:measurement_model}
\end{align}
Optimal alignment of translation invariant measurements $\mathbf{s}_{ji}$ and $\mathbf{o}_{ji}$ is given by the state $\mathbf{x}$, that minimizes Equation \eqref{eq:measurement_model}. We force the pseudo measurement model $\mathbf{z}^h=0$. The pseudo measurements is associated with uncertainties that depend on $\mathbf{x}_t, \mathbf{s}_{ji}$ and $\mathbf{o}_{ji}$.
We assume that $\mathbf{x}$ and $\mathbf{z}_t$ are Gaussian distributed. Subsequently, by considering a static process model, the Kalman equations are given by
\begin{align}
    \mathbf{x}_{t} &= \bar{\mathbf{x}}_{t-1} - \mathbf{K}_t \left( \mathbf{H}_t \bar{\mathbf{x}}_{t-1} \right) \\
    \Sigma^{\mathbf{x}}_{t} &= \left( \mathbf{I} - \mathbf{K}_t \mathbf{H}_t \right) \bar{\Sigma}^{\mathbf{x}}_{t-1} \\
    \mathbf{K}_t &= \bar{\Sigma}^\mathbf{x}_{t-1} \mathbf{H}_t^T \left( \mathbf{H}_t\bar{\Sigma}^\mathbf{x}_{t-1} \mathbf{H}_t^T + \Sigma_t^{\mathbf{h}}\right)^{-1} \quad , \label{eq:kalman_equations}
\end{align}
where $\bar{\mathbf{x}}_{t-1}$ is the normalized mean of the state estimate at $t-1$, $\mathbf{K}_t$ is the Kalman gain and $\bar{\Sigma}^{\mathbf{x}}_{t-1}$ is the covariance matrix of the state at $t-1$. The parameter $\Sigma_t^{\mathbf{h}}$ is the measurement uncertainty at timestep $t$ which is state-dependent and is defined as follows~\cite{choukroun2006novel}:
\begin{align}
    \Sigma_t^{\mathbf{h}} = \frac{1}{4}\rho\left[ tr(\bar{\mathbf{x}}_{t-1}\bar{\mathbf{x}}_{t-1}^T + \bar{\Sigma}^{x}_{t-1})\mathbb{I}_4 - (\bar{\mathbf{x}}_{t-1}\bar{\mathbf{x}}_{t-1}^T + \bar{\Sigma}^{x}_{t-1} )\right] \quad ,
    \label{eq:choukron}
\end{align}
where $\rho$ is a constant which corresponds to the uncertainty of the correspondence measurements and is set empirically.
In order for the state to represent a rotation, a common technique is used to normalize the state after a prediction step as
\begin{equation}
    \bar{\mathbf{x}}_{t} = \frac{\mathbf{x}_{t}}{||\mathbf{x}_{t}||_2} \quad \bar{\Sigma}^{\mathbf{x}}_{t} = \frac{\Sigma^{\mathbf{x}}_{t}}{||\mathbf{x}_{t}||_2^2} \quad .
\end{equation}
Once the rotation is estimated using the Kalman Filter, computing the translation from~\eqref{eq:generativemodel} as:
\begin{equation}
    \mathbf{t} = \frac{\sum_{i=0}^{N} \mathbf{s}_i}{N} - \mathbf{R}\frac{\sum_{i=0}^{N} \mathbf{o}_i}{N} \quad .
\end{equation}
With each iteration of the update step of the Kalman filter, we obtain a new homogeneous transformation ${}^{W} H_{\mathcal{F}}$ which is then used to transform the model. The transformed model is used to recompute correspondences and repeat the Kalman Filter update steps. Similar to ICP~\cite{besl1992method}, we calculate the change in homogeneous transformation between iterations and/or maximum number of iterations in order to check for convergence.

\subsection{Next Best Touch Selection}
\label{sec:active_selection}

\begin{table*}[t!]
\caption{Divergence/ distance measures for multivariate Gaussian distributions $p_i = \mathcal{N}(\mu_i, \Sigma_i)$ and $p_j = \mathcal{N}(\mu_j, \Sigma_j)$}
\label{tab:eq}
\centering
\resizebox{0.85\textwidth}{!}{%
\begin{tabular}{@{}lll@{}}
\toprule 
\multicolumn{1}{c}{Name} & \multicolumn{1}{c}{\textbf{$D(p_{i}||p_j)$}} & \multicolumn{1}{c}{Comments} \\ \midrule
Kullback-Leibler divergence                                    &   $\frac{1}{2}[ log\frac{|({\Sigma}_{j})|}{|({{\Sigma}}_{i})|} + tr({\Sigma}_{j}^{-1} {{\Sigma}}_{i})) - d + ({{\mu}_{i}} - {\mu}_{j})' {\Sigma}_{j}^{-1} ({{\mu}}_{i} - {\mu}_{j})]$ &               d = 4 in our case                                                       \\
R\'enyi divergence                                          &         $ \frac{\alpha}{2}(\mu_i - \mu_j)'(\Sigma_\alpha)^*(\mu_i - \mu_j) - \frac{1}{2(\alpha-1)}log \frac{|(\Sigma_\alpha)^*|}{|\Sigma_i^{1-\alpha}||\Sigma_j^{\alpha}|} $
& $(\Sigma_{\alpha})^* = \alpha\Sigma_j + (1-\alpha)\Sigma_i$
\\
Fisher Information metric                                        &        $|\Sigma_j^{-1}(\mu_i - \mu_j)|^2 + tr(\Sigma_j^{-2}\Sigma_i -2\Sigma_j^{-1} + \Sigma_i^{-1})$                                                                 &\\
Bhattacharya distance                                     &       $\frac{1}{8}(\mu_i - \mu_j)'\Sigma^{-1}(\mu_i -\mu_j) + \frac{1}{2} log (\frac{|\Sigma|}{\sqrt{|\Sigma_i||\Sigma_j|}})$                                                                  & $\Sigma = \frac{\Sigma_i + \Sigma_j}{2}$\\
2-Wasserstein distance$^2$                                      &       $ |(\mu_i - \mu_j)|^2 + tr(\Sigma_i + \Sigma_j -2\sqrt{\sqrt{\Sigma_i}\Sigma_j\sqrt{\Sigma_i}}))$                                                              &    \\ \bottomrule
\end{tabular}
}
\end{table*}

To reduce the number of touches required to converge to the true position of the object, we need to make an informed decision on which action $\mathbf{a}_{t}$ to perform next based on the current state estimate. We generate the set of actions $\mathcal{A}$ by sampling uniformly along the faces of a bounding box on the current estimate of the object pose. We define an action as a ray represented by a tuple $\mathbf{a} = (\mathbf{n}, \mathbf{d})$, with $\mathbf{n}$ as the start point and $\mathbf{d}$ the direction of the ray. We seek to choose the action $\mathbf{a}^{*}_{t}$, that \textit{maximizes} the overall \textit{Information Gain}.
However, as the future measurements $\mathbf{z}_{t}$ are hypothetical, we approximate our action-measurement model $p(\mathbf{z}_{t} | \mathbf{x}, \mathbf{a}_{t})$ as a ray-mesh intersection in simulation to extract the hypothetical measurement given a certain action when the object is at the estimated pose.
For each hypothetical action $\hat{\mathbf{a}}_t \in \mathcal{A}(\mathbf{x}_{t-1})$ and the hypothetical measurement $\hat{\mathbf{z}}_{t}$, we estimate the posterior by $p(\mathbf{x} | \hat{\mathbf{z}}_{1:t}, \hat{\mathbf{a}}_{1:t})$ known as the \textit{one-step look ahead}. In order to calculate the information gain, we empirically compare a few widely used information theoretic criteria as follows:
\begin{enumerate}
    \item \textit{Kullback–Leibler divergence (KL)}~\cite{kullback1951information} (or relative entropy) measures the how different one probability distribution is from another. For two discrete probability distributions $p_i$ and $p_j$ defined in the probability space $s\in \mathcal{S}$, $D_{KL}(p_i || p_j) = \sum_{s\in\mathcal{S}} p_i(s) log\frac{p_i(s)}{p_j(s)}$.  
    \item \textit{R\'enyi divergence}\cite{renyi1961measures} generalises the KL divergence and is defined as: $D_{\alpha}(p_i || p_j) = \frac{1}{1 - \alpha}log(\sum_{s\in\mathcal{S}}\frac{p_i^{\alpha}(s)}{p_j^{\alpha-1}(s)})$ for $0 < \alpha < \infty$ and $\alpha \neq 1$. For limiting case $\alpha \rightarrow 1$, the R\'enyi divergence is the same as KL divergence.
    \item \textit{Fisher information metric}\cite{fisher1922mathematical} measures the amount of information that an observable random variable $X$ carries about an unknown parameter $\theta$ upon which the probability of $X$ depends. It is defined as second derivative of the KL divergence.
    \item \textit{Bhattacharya distance}~\cite{bhattacharyya1943measure} measures the relative closeness of two probability distributions. It is defined as $D_B(p_i || p_j) = -\ln(\sum_{s\in\mathcal{S}}\sqrt{p_i(s)p_j(s)})$.
    \item \textit{Wasserstein distance}~\cite{olkin1982distance} (or earth mover distance) is a way to compare two probability distributions, where one distribution is derived from the other by small, non-uniform perturbations (random or deterministic). It is defined as $W_p(\lambda,\nu) = (inf(\mathbb{E}[d(X,Y)^p])^{1/p}$ for two distributions $\lambda, \nu$ and the infimum is taken over all joint distributions of the random variables $X$ and $Y$ with marginals $\lambda$  and $\nu$  respectively~\cite{enwiki}.
\end{enumerate}
Given that the prior and posterior are multivariate Gaussian distributions, we have closed form solutions for each of the divergence or distance metrics as described in Table~\ref{tab:eq}. 
Therefore we perform the most optimal action $\mathbf{a}_{t}^*$ given by
\begin{align}
    &\mathbf{a}_{t}^* 
    &= \argmax_{\hat{\mathbf{a}}_{t}} D({p(\mathbf{x} | \hat{\mathbf{z}}_{1:t},  \hat{\mathbf{a}}_{1:t})}||{p(\mathbf{x} | \mathbf{z}_{1:t-1}, \mathbf{a}_{1:t-1})} \quad . \label{eq:kl_div_prio_post}
\end{align}

\section{EXPERIMENTS}
\label{sec:experiment}
We perform simulation experiments on the Bunny dataset of the Stanford Scanning Repository for calculating the next best touch using various information gain criteria. 
We added noise that is sampled randomnly from a normal distribution $\mathcal{N}(0, 5\times10^{-3})$ to the cloud obtained from the meshes, henceforth called \textit{scene}. We set the initial start pose for each model sampled uniformly from $[-50, 50]mm$ and $[-30^o, 30^o]$ for position and orientation respectively. The initial state $\mathbf{x}_0$ is obtained from the initial start pose and the initial covariance $\Sigma^{\mathbf{x}}_0$ is set to $10^4*\mathbb{I}_4$.
A touch action is represented by a point measurement derived from the ray-mesh intersection as described in Section~\ref{sec:methods}.
In order to initialise the TIQF, we sample 3 measurements by random touches and begin the registration process with the 4th touch.
In particular, we used an $\alpha =0.3$ for R\'enyi divergence which was empirically tuned.
All simulation were executed on a workstation running Ubuntu 18.04 with 8 core Intel i7-8550U CPU @ 1.80GHz and 16 GB RAM.
We report the simulation results for 6 repeated runs on each metric and show the average $L_2$ norm of the absolute error in position (m), rotation ($^{o}$) and Average Distance of model points with Indistinguishable views (ADI) metric~\cite{hinterstoisser2012model}. The ADI metric is defined as the matching score between the estimated pose and the ground truth pose for all points $p_1, p_2 \in \mathcal{O} = \{1,2,\dots, M\}$.  
\begin{equation}
    ADI = \frac{1}{M}\sum_{\mathbf{p}_1 \in \mathcal{O}} \min_{\mathbf{p}_2 \in \mathcal{O}} || (\mathbf{R}_{gt}\mathbf{p}_1 + \mathbf{t}_{gt}) - (\hat{\mathbf{R}}\mathbf{p}_2 + \hat{\mathbf{t}}) || \quad ,
    \label{eq:}
\end{equation}

\begin{figure*}[t!]
    \centering
    \includegraphics[width = \textwidth, height = 5cm]{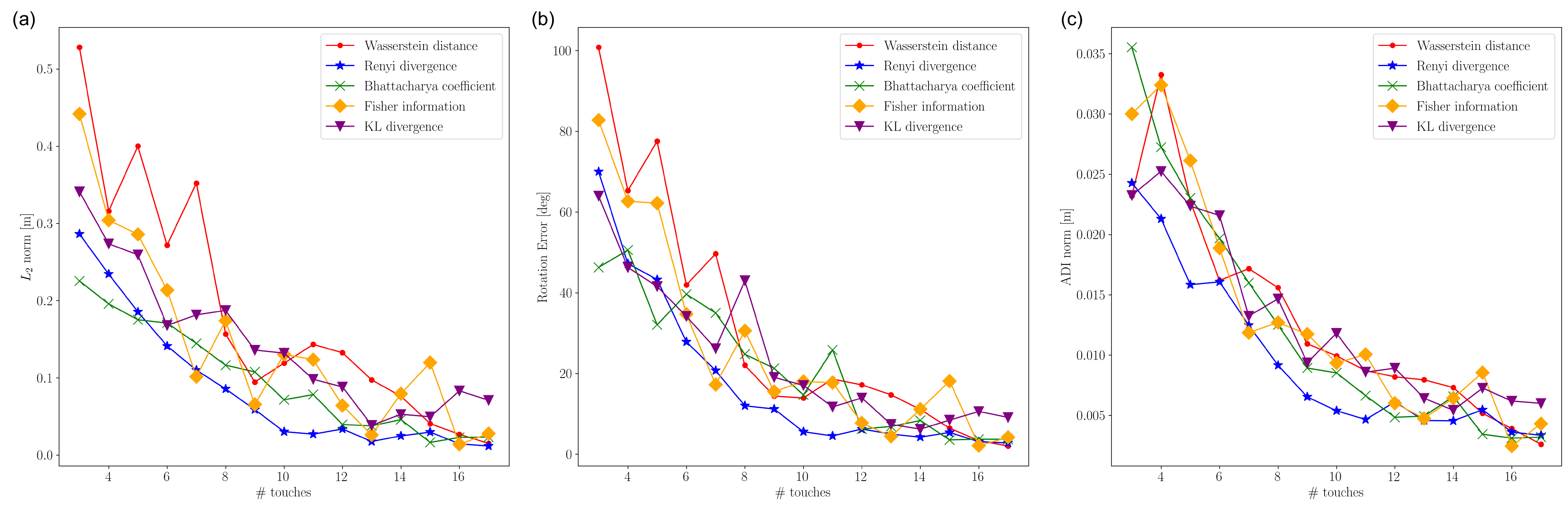}
    \caption{Simulation results on the Stanford Scanning Bunny dataset for the various information gain metrics: (a) Average $L_2$ norm of the position error with number of touch measurements, (b) average $L_2$ norm of the rotation error with number of touch measurements, (c) average ADI with number of touch measurements for 6 repeated runs.}
    \label{fig:plots}
\end{figure*}

\begin{figure*}[t!]
    \centering
    \includegraphics[width = \textwidth]{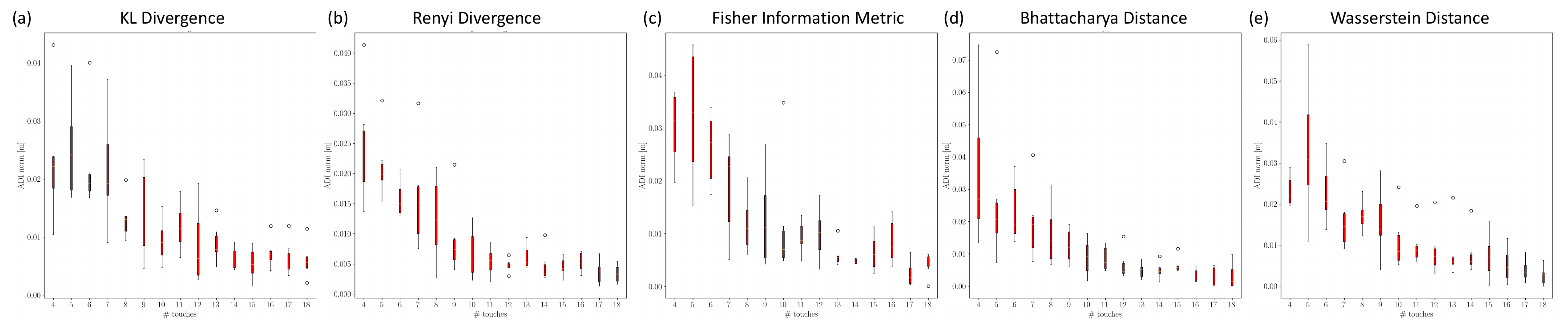}
    \caption{Box-and-whisker plots of the ADI metric for each criteria with increasing number of measurements.}
    \label{fig:plots_std}
\end{figure*}
The results of the experiments are presented in the Figure~\ref{fig:plots} and Figure~\ref{fig:plots_std}.
Across all the selected criteria, the pose error iteratively reduces with increasing number of measurements that are actively selected. In fact, for all the criteria the ADI metric is $<1$cm for $\sim 15$ measurements. Hence it shows that the active touch strategy using any information gain criteria with our proposed pose estimation approach helps to reduce the pose error with few measurements. We note very similar performance for each criteria and it is not straightforward to conclude if any particular criteria is better. In fact we see comparatively lower variance for R\'enyi divergence and Wasserstein distance. In terms of accuracy, we note that KL divergence has comparatively slightly lower accuracy with other criteria however still within the acceptable accuracy range of $<1$cm.   
Hence, we provide initial empirical evaluation in simulation showing the adaptability of our proposed TIQF pose estimation approach with various information gain criteria. As future work, we will investigate the same with real world data collected from tactile sensors.

\section{CONCLUSIONS}
\label{sec:conclusions}
In this paper, we empirically evaluated various information gain criteria for action selection in the context of object pose estimation. Due to our novel TIQF formulation enforcing Gaussian distributions for prior and posterior rotation estimate, all criteria have closed form solutions that allows to exhaustively search for the next best action with marginal computation time overhead. We demonstrated similar performance in terms of pose accuracy with sparse measurements ($<15$ points) across all the selected criteria. This shows the adaptability and effectiveness of our probabilistic pose estimation method for various information gain criteria. Finally, this work also provides the theoretical framework for employing various well known and uncommon information theoretic criteria for action selection.



\section*{ACKNOWLEDGMENT}
We would like to sincerely thank Mr. Michael Gentner for assistance with the figures. 

\bibliography{IEEEexample}
\bibliographystyle{IEEEtran}

\end{document}